\pdfoutput=1
\documentclass[11pt]{article}

\usepackage[final]{acl}

\usepackage{times}
\usepackage{latexsym}

\usepackage{algorithm}
\usepackage{algorithmic}
\usepackage{booktabs}
\usepackage{amsmath}
\usepackage{multirow}
\usepackage{makecell}
\usepackage{graphicx}
\usepackage{subcaption}
\usepackage{listings}
\usepackage{subcaption}
\usepackage{enumitem}
\usepackage[T1]{fontenc}
\usepackage[utf8]{inputenc}

\usepackage{microtype}

\usepackage{inconsolata}

\usepackage{graphicx}

\title{AgentCourt: Simulating Court with Adversarial Evolvable Lawyer Agents}
\title{AgentCourt: Simulating Court with Adversarial Evolvable Lawyer Agents}

\author{
 \textbf{Guhong Chen\textsuperscript{1,2,3,}}\thanks{Equal contribution},
 \textbf{Liyang Fan\textsuperscript{2,}}\footnotemark[1],
 \textbf{Zihan Gong\textsuperscript{1,2,}}\footnotemark[1],
 \textbf{Nan Xie\textsuperscript{2}},
 \textbf{Zixuan Li\textsuperscript{2}},
\\
 \textbf{Ziqiang Liu\textsuperscript{2}},
 \textbf{Chengming Li\textsuperscript{4}},
 \textbf{Qiang Qu\textsuperscript{2}},
 \textbf{Hamid Alinejad-Rokny\textsuperscript{5}},\\
 \textbf{Shiwen Ni\textsuperscript{2,3,}}\thanks{Min Yang and Shiwen Ni are corresponding authors.},
 \textbf{Min Yang\textsuperscript{2,3,6,}}\footnotemark[2]
\\
 \textsuperscript{1}Southern University of Science and Technology\\
 \textsuperscript{2}Shenzhen Key Laboratory for High Performance Data Mining, \\Shenzhen Institutes of Advanced Technology, Chinese Academy of Sciences\\
 \textsuperscript{3}SIAT-DELI AI and Law Joint Lab~
 \textsuperscript{4}Shenzhen MSU-BIT University~
 \textsuperscript{5}UNSW Sydney~
 \textsuperscript{6}SUAT
\\
 \texttt{\{gh.chen2, sw.ni, min.yang\}@siat.ac.cn}
}

\begin{document}
\maketitle
\begin{abstract}
Current research in LLM-based simulation systems lacks comprehensive solutions for modeling real-world court proceedings, while existing legal language models struggle with dynamic courtroom interactions. We present \textbf{AgentCourt}, a comprehensive legal simulation framework that addresses these challenges through adversarial evolution of LLM-based agents. Our AgentCourt introduces a new adversarial evolutionary approach for agents called \textbf{AdvEvol}, which performs dynamic knowledge learning and evolution through structured adversarial interactions in a simulated courtroom program, breaking the limitations of the traditional reliance on static knowledge bases or manual annotations.
%Our system innovatively combines three specialized knowledge bases—legal provisions, practical experience, and case precedents—with an automated knowledge evolution mechanism that enables continuous learning through simulated court proceedings. %The core of our approach is the \textbf{Adv}ersarial \textbf{Evol}ution (AdvEvol) method, which allows legal agents to acquire and refine their capabilities through structured courtroom debates. 
By simulating 1,000 civil cases, we construct an evolving knowledge base that enhances the agents' legal reasoning abilities. The evolved lawyer agents demonstrated outstanding performance on our newly introduced \textbf{CourtBench} benchmark, achieving a 12.1\% improvement in performance compared to the original lawyer agents. Evaluations by professional lawyers confirm the effectiveness of our approach across three critical dimensions: cognitive agility, professional knowledge, and logical rigor. Beyond outperforming specialized legal models in interactive reasoning tasks, our findings emphasize the importance of adversarial learning in legal AI and suggest promising directions for extending simulation-based legal reasoning to broader judicial and regulatory contexts\footnote{https://github.com/relic-yuexi/AgentCourt}. %Our anonymized datasets, implementation, and demonstration videos are provided in the supplementary materials and will be made publicly available upon publication.
\end{abstract}

\section{Introduction}
Large language models (LLMs) have shown remarkable success in simulating real-world professional scenarios, from medical consultations to educational interactions \cite{li2024agent}. However, in the legal domain, comprehensive simulation of court proceedings remains an underexplored challenge. While existing legal language models excel at static tasks such as legal provision retrieval and question answering \cite{lai2023largelanguagemodelslaw}, they struggle with dynamic courtroom interactions. For instance, these models can accurately recite Articles of Civil Law and regulations but often fail to leverage them effectively in adversarial court debates. More critically, models like ChatLaw-33B \cite{cui2024chatlaw,ChatLaw} exhibit severe overfitting to standardized legal tasks, sometimes losing the ability to generate coherent responses in interactive courtroom scenarios.

\begin{figure*}[ht]
\centering
\includegraphics[width=0.99\linewidth]{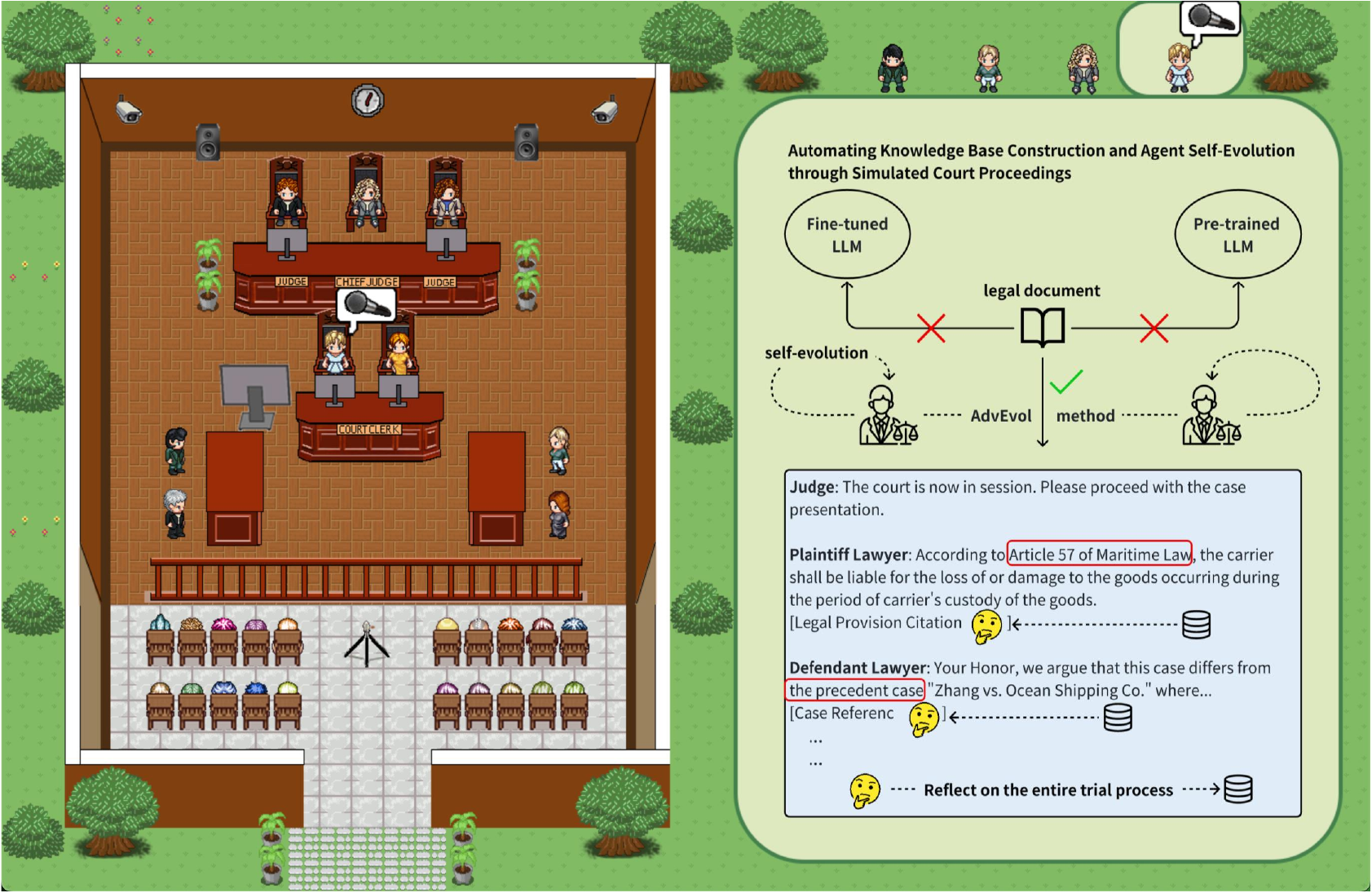}
\caption{(Left) The mock courtroom sandbox interface supporting character movement and real-time dialogue, with a complete case demonstration available in the supplementary materials. (Right) The automated knowledge base construction and self-evolution of lawyer agent capabilities through the mock courtroom. The red boxes highlight key components corresponding to Formula \eqref{eq1} and Formula \eqref{eq7} in Section~\ref{sec:AdvEvol}, which utilize knowledge from previous cases to assist in answering questions and enable continuous learning through post-trial reflection.}
\label{fig:scene}
\end{figure*}

To address these limitations, we present AgentCourt, an innovative framework for simulating civil court proceedings through LLM-based agents. Unlike previous approaches that focus on specific legal tasks, our system creates a complete courtroom environment where multiple agents—including judges, attorneys, and other participants—engage in structured legal discourse, as shown in Figure \ref{fig:scene}. At its core, AgentCourt employs an \textbf{Adv}ersarial \textbf{Evol}ution (AdvEvol) method that enables continuous knowledge acquisition through simulated court interactions, eliminating the need for extensive manual annotation, fine-tuning, or specialized legal pre-training.

A key innovation of our approach is its automated knowledge evolution mechanism. By simulating court proceedings, lawyer agents construct and refine three specialized knowledge bases: a legal provisions memory for statutory understanding, an experience base for debate strategies, and a case library for precedent analysis. Starting with only complaints and defense statements from real cases, agents engage in adversarial debates to autonomously build and evolve their legal knowledge. This multi-faceted knowledge structure, combined with our adversarial learning strategy, enables agents to develop sophisticated legal reasoning capabilities that extend beyond simple information retrieval or pattern matching.

Through the simulation of 1,000 civil cases, we demonstrate significant improvements in the legal capabilities of our agents. Our evolved agents achieve performance comparable to GPT-4o on dynamic courtroom tasks while significantly outperforming specialized legal models. A particularly noteworthy finding is the contrast in model behavior: while existing legal models like ChatLaw-33B perform well on standardized tasks, they struggle significantly with dynamic courtroom dialogue, often failing to generate valid responses.

To facilitate systematic evaluation of such capabilities, we introduce CourtBench, a dedicated benchmark designed to assess interactive legal reasoning. Our findings underscore the importance of interactive learning in developing robust legal AI systems capable of handling dynamic legal scenarios.

The main contributions of our work include:

\begin{itemize} \item We propose \textbf{AgentCourt}, the first court simulation framework enabling multi-party legal interactions and complex reasoning through the adversarial evolution of LLM-based agents.

\item A novel automated knowledge evolution mechanism that requires only real-world complaints and defense statements as initial input. Through self-play court debates, agents autonomously construct and refine legal expertise across three specialized knowledge bases: a legal provisions memory for statutory understanding, an experience base for debate strategies, and a case library for precedent analysis. This self-evolving approach facilitates continuous expansion of knowledge without manual annotation, offering a scalable solution for future acquisition of legal knowledge.

\item We propose \textbf{CourtBench}, a newly introduced benchmark designed to evaluate models’ capabilities in dynamic courtroom dialogue, addressing a critical gap in legal AI evaluation and ensuring systematic assessment of interactive legal reasoning.
\end{itemize}

\section{Related Work}
\subsection{LLMs in the Legal Domain}
\label{sec:llms_legal}

AI applications in the legal domain have progressed significantly, particularly with the development of large language models (LLMs). These models have demonstrated strong potential in various legal tasks, including case prediction, legal research, and document analysis \cite{lai2023largelanguagemodelslaw, hamilton2023blindjudgementagentbasedsupreme}. Recent studies have explored various strategies to enhance LLMs' legal reasoning capabilities, leading to the emergence of several specialized legal models.

For instance, Lawyer-LLaMA-13B \cite{huang2023lawyer}, a 13B-parameter model fine-tuned on Chinese legal documents, has shown promising results in legal consultation. HanFei-7B \cite{HanFei} focuses on legal knowledge representation and statutory interpretation, while ChatLaw-33B \cite{cui2024chatlaw,ChatLaw} employs a mixture-of-experts architecture integrated with a legal knowledge graph to improve reasoning capabilities. Other approaches, such as DISC-LawLLM, highlight the effectiveness of fine-tuned LLMs in delivering intelligent legal services \cite{yue2023disclawllmfinetuninglargelanguage}. The PLJP framework, on the other hand, enhances case judgment prediction accuracy by combining LLMs with domain-specific models \cite{wu2023precedentenhancedlegaljudgmentprediction}. Similarly, DeliLaw \cite{xie2024delilaw} has demonstrated efficiency in handling legal inquiries through a dialogue-based system.

Despite these advancements, existing legal AI models remain largely confined to static, well-defined tasks, struggling to handle dynamic legal interactions. Although these models are trained on extensive legal corpora and leverage sophisticated architectures, they continue to face limitations when addressing complex legal queries and simulating real-world court proceedings. Many of these systems, including those mentioned above, remain task-specific and struggle with fully replicating the legal reasoning process and facilitating multi-party interactions \cite{janatian2023textstructureusinglarge, jin2023understandlegaldocumentscontextualized}.
\subsection{LLMs for Real-World Simulation}

LLM-based multi-agent systems are a rapidly advancing area of AI research, leveraging collaborative agents to solve complex problems. These systems excel in combining knowledge sharing, cognitive synergy, and decision-making improvements \cite{talebirad2023multiagent, handler2023balancing}.

The potential of multi-agent LLM-based systems has been demonstrated in various domains. In natural language processing, they have improved language understanding and generation tasks \cite{tan2023large}. In robotics, they have enhanced decision-making in human-robot interactions \cite{kim2024understanding}. Similarly, task planning and execution have benefited from multi-agent approaches, enabling the decomposition and collaborative completion of complex tasks \cite{yang2024embodied}. The education sector has leveraged these systems for personalized learning experiences and intelligent tutoring \cite{yin2023lumos}. In finance, LLM-based agents contribute to market analysis, risk assessment, and investment decision-making \cite{nascimento2023self}.

A particularly relevant example is Agent Hospital \cite{li2024agent}, a simulation framework that models a hospital environment using autonomous agents representing doctors, nurses, and patients. The system includes comprehensive disease treatment simulations, autonomous learning without manual annotation, and state-of-the-art medical performance benchmarks. Agent Hospital highlights the effectiveness of multi-agent LLMs in complex, specialized domains, showcasing their potential for professional training and decision support.

Building on these advancements, our AgentCourt extends the multi-agent approach to the legal domain while addressing the shortcomings of current legal AI systems. By simulating a civil court environment, AgentCourt provides comprehensive legal scenario simulations, incorporating both the dynamic nature of courtroom interactions and automated construction of knowledge bases through simulation. This approach not only bridges a critical gap in legal AI research but also demonstrates the broader potential of multi-agent systems in advancing professional domain simulations.

\section{Court Simulation}
\subsection{Agent Design}
We design an agent framework that simulates real litigation scenarios, incorporating both core legal agents and auxiliary agents. Each agent is built upon GPT-4o-mini and optimized for specific legal roles. The detailed prompt templates are provided in Appendix~\ref{appendix:Agents and Prompts}.

The core legal agents consist of two lawyer agents and one judge agent. The lawyer agents dynamically assume plaintiff or defendant roles, accumulating experience from different litigation perspectives through a bidirectional learning mechanism. They are responsible for case analysis, evidence organization, and courtroom debates. The judge agent oversees trial proceedings, ensures procedural adherence, extracts key dispute points, and delivers final judgments.

To enhance the realism and completeness of the simulation, we introduce auxiliary agents, including a clerk, a plaintiff, and a defendant. The clerk agent manages procedural progression and maintains trial documentation, while the plaintiff and defendant agents provide essential case information. Together, these agents create a fully functional litigation ecosystem. The design of the agent roles is illustrated in Figure~\ref{fig:agent} in Appendix~\ref{appendix:Agent Visualization}.

Agent interactions can be formalized as:
\[
I(a_i, a_j, t) = f_{interact}(D_L(s_t), D_O(s_t))
\]
where $I$ represents the interaction result, $a_i$ and $a_j$ are the interacting agent pair at time step $t$.

Agent decision mechanisms can be formalized as:
\[
D_L(s_t) = f_{LLM}(s_t, \mathcal{K}_t)
\]
\[
D_O(s_t) = f_{LLM}(s_t)
\]
where $D_L$ denotes the lawyer agent's decision function, which depends on the current state $s_t$ and the knowledge base $\mathcal{K}_t$. Similarly, $D_O$ represents the decision functions of other agents (e.g., judge, clerk), which rely solely on the current state. The function $f_{LLM}$ encapsulates the language model's fundamental reasoning capabilities. This design allows lawyer agents to utilize accumulated knowledge to enhance decision-making while ensuring stable functionality for other agent roles.

\subsection{Simulation Workflow}
The simulation workflow comprises three main phases: pre-trial preparation, court proceedings, and knowledge construction. During pre-trial preparation, we curate an experimental dataset by processing 1,000 real civil cases from major Chinese courts (2018–2020), with detailed data processing described in Appendix~\ref{appendix:Data Settings and Processing}.

The court proceedings adhere to standard civil court procedures, beginning with the clerk's announcement and the judge's validation, followed by case presentation and structured debates between lawyers. The detailed court protocol and interaction patterns are outlined in Appendix~\ref{appendix:Simulation Workflow Details}.

In the knowledge construction phase, lawyer agents reflect on court sessions to refine their legal reasoning capabilities through our AdvEvol method, which is detailed in Section~\ref{sec:AdvEvol}. The complete simulation workflow is illustrated in Appendix~\ref{appendix:System Overview}, Figure~\ref{fig:overall process}.

\subsection{AdvEvol Method}
\label{sec:AdvEvol}
To enhance the legal reasoning capabilities of simulated agents, we introduce the Adversarial Evolution (AdvEvol) method, a novel approach that diverges fundamentally from existing legal AI systems. Traditional methods predominantly rely on static knowledge bases or manual annotations, constraining their adaptability to diverse legal scenarios. In contrast, AdvEvol facilitates dynamic knowledge acquisition through structured adversarial interactions within simulated court proceedings.

The core innovation of our method lies in its three synergistic knowledge bases:
\[
\mathcal{K} = \{\mathcal{R}, \mathcal{E}, \mathcal{C}\}
\]
where $\mathcal{R}$ represents the regulations memory for legal provisions, $\mathcal{E}$ denotes the experience base for debate strategies, and $\mathcal{C}$ corresponds to the case library for precedent analysis.

Previous studies, such as AI-town~\cite{park2023generative} and MedAgent-Zero~\cite{li2024agent}, have primarily focused on cooperative agent communication. In contrast, our approach utilizes adversarial interactions within court simulations to facilitate more targeted and effective knowledge evolution. The knowledge acquisition process is formalized as:
\[
\mathcal{K}_{t+1} = f_{evolve}(\mathcal{K}_t, \mathcal{G}_t)
\]
where $\mathcal{G}_t$ denotes the dialogue history at time $t$, and $f_{evolve}$ encapsulates our three-tier evolution strategy, detailed in the following sections.

\subsubsection{Regulations Memory Shaping}
Legal provisions form the foundation of judicial reasoning and decision-making. The regulations memory $\mathcal{R}$ systematically captures and organizes legal provisions through continuous learning during court proceedings, ensuring agents maintain a comprehensive understanding of applicable laws. The system actively identifies and extracts explicitly referenced legal provisions:
\begin{equation}
    \mathcal{R}_{direct} = f_{extract}(\mathcal{G})
    \label{eq1}
\end{equation}
while also analyzing case contexts to identify potentially relevant provisions:
\begin{equation}
    \mathcal{R}_{reflect} = f_{reflect}(\mathcal{G})
\end{equation}
The knowledge base is continuously refined through:
\begin{equation}
    \mathcal{R}_{t+1} = f_{refine}(\mathcal{R}_t, \mathcal{R}_{direct}, \mathcal{R}_{reflect})
\end{equation}

\subsubsection{Experience Base Expansion}
The experience base $\mathcal{E}$ serves as a repository of legal expertise, integrating self-reflective insights and opponent-learning experiences to enhance agents' legal reasoning and strategic decision-making abilities. The self-reflection component processes case experiences as:
\begin{equation}
    \mathcal{E}_{self} = f_{reflect}(agent_i, \mathcal{G}, \mathcal{R})
\end{equation}
This mechanism ensures coherent legal arguments by analyzing case backgrounds, dispute focal points, and strategic approaches, enabling accumulated experience to contribute to sophisticated legal reasoning.

The adversarial learning component extracts insights from opponent strategies:
\begin{equation}
    \mathcal{E}_{adv} = f_{observe}(agent_i, agent_j, \mathcal{G}, key)
\end{equation}
focusing on legal provision selection, argument coherence, and expression effectiveness. The experience base evolves iteratively through:
\begin{equation}
    \mathcal{E}_{t+1} = f_{refine}(\mathcal{E}_t, \mathcal{E}_{self}, \mathcal{E}_{adv})
\end{equation}

\subsubsection{Case Library Construction}
The case library $\mathcal{C}$ transforms historical cases into structured knowledge representations. During knowledge extraction, the system performs analysis as:
\begin{equation}
    c_{refined} = f_{distill}(\mathcal{G}, key, \mathcal{R})
    \label{eq7}
\end{equation}
This process extracts key elements from cases, including case background, type, keywords, quick reaction points, and response directions. The structured representation is defined as:
\begin{equation}
    \mathcal{C}_{structured} = \{(c, t, k, r, d)\}
\end{equation}
where $c$ contains the case name and background description, $t$ denotes the case category (e.g., labor dispute, contract dispute), $k$ includes $3\sim5$ essential terms, $r$ stores quick response points, and $d$ contains potential response strategies. This structured format supports efficient case retrieval and knowledge application by legal agents.

The case library is updated dynamically through:
\begin{equation}
    \mathcal{C}_{t+1} = f_{refine}(\mathcal{C}_t, c_{new})
\end{equation}

Based on these mechanisms, we summarize our complete framework in Algorithm~\ref{alg:agentcourt}, which provides a high-level overview of the AgentCourt simulation process and its knowledge evolution procedure.

\begin{algorithm}[t]
\small
\caption{AgentCourt Framework with AdvEvol Knowledge Evolution}
\label{alg:agentcourt}
\renewcommand{\algorithmicrequire}{\textbf{Input:}}
\renewcommand{\algorithmicensure}{\textbf{Output:}}
\begin{algorithmic}[1]
\REQUIRE Complaint statements $\mathcal{S}_c$, Defense statements $\mathcal{S}_d$
\ENSURE Evolved knowledge bases $\mathcal{K}$ \& Enhanced lawyer agents
\STATE Initialize knowledge bases $\mathcal{R}, \mathcal{E}, \mathcal{C} \leftarrow \emptyset$
\STATE Initialize agents $\mathcal{A} \leftarrow$ \{judge, plaintiff lawyer, defendant\\ \hspace*{4em} lawyer, ...\}
\FOR{each case $(s_c, s_d)$ in $(\mathcal{S}_c, \mathcal{S}_d)$}
    \STATE $\textit{context} \leftarrow \text{ProcessCase}(s_c, s_d)$
    \WHILE{not session\_complete}
        \STATE $\textit{current\_agent} \leftarrow \text{SelectAgent}(\mathcal{A})$
        \IF{$\textit{current\_agent} = \text{lawyer}$}
            \STATE $\textit{response} \leftarrow \text{GenerateResponse}(\textit{current\_agent},$ \\ \hspace*{4em}$\textit{context}, \mathcal{K})$
        \ELSE
            \STATE $\textit{response} \leftarrow \text{GenerateResponse}(\textit{current\_agent},$ \\ \hspace*{4em}$\textit{context})$
        \ENDIF
        \STATE $\textit{context} \leftarrow \text{UpdateContext}(\textit{context}, \textit{response})$
    \ENDWHILE
    \STATE \textcolor{gray}{// Knowledge Evolution after judge's final verdict}
    \STATE $\mathcal{R}_{t+1} \leftarrow f_{\text{refine}}(\mathcal{R}_t, \mathcal{R}_{\text{direct}}, \mathcal{R}_{\text{reflect}})$ \hfill \textcolor{gray}{// Eq.(3)}
    \STATE $\mathcal{E}_{t+1} \leftarrow f_{\text{refine}}(\mathcal{E}_t, \mathcal{E}_{\text{self}}, \mathcal{E}_{\text{adv}})$ \hfill \textcolor{gray}{// Eq.(6)}
    \STATE $\mathcal{C}_{t+1} \leftarrow f_{\text{refine}}(\mathcal{C}_t, c_{\text{new}})$ \hfill \textcolor{gray}{// Eq.(9)}
    \STATE $\mathcal{K} \leftarrow \{\mathcal{R}_{t+1}, \mathcal{E}_{t+1}, \mathcal{C}_{t+1}\}$
\ENDFOR
\RETURN $\mathcal{K}$ \& Enhanced lawyer agents
\end{algorithmic}
\end{algorithm}

\section{Experiments}
\subsection{Experimental Setup}
We evaluate our approach through an integrated framework combining simulated courtroom debates and benchmark assessments. The simulated trial environment is constructed using authentic legal documents from actual civil cases, including real-world complaints and defense briefs. Within this environment, models alternately assume the roles of plaintiff and defendant counsel. Their performance is systematically assessed across three empirically validated dimensions of legal expertise identified through consultations with legal practitioners:
\begin{itemize}
    \item \textbf{Cognitive Agility:} The ability to comprehend and respond to opposing arguments, identify weaknesses, and effectively integrate information for counterarguments.
    
    \item \textbf{Professional Knowledge:} Legal expertise, measured through accurate citation of laws and precedents, understanding of legal principles, and clear articulation of arguments.
    
    \item \textbf{Logical Rigor:} The consistency and coherence of argumentation, including structural clarity and logical reasoning.
\end{itemize}

Our evaluation framework combines expert human assessment with automated LLM evaluation. Legal professionals from a leading law firm assess 40 distinct cases (20 cases per role) spanning four major civil dispute categories: contract disputes, tort cases, marriage \& family cases, and property rights cases (5 cases each). For automated evaluation, we employ GPT-4o-mini, applying identical criteria to ensure consistent comparative analysis. Both human and LLM evaluations adopt a win-tie-loss framework across all three dimensions. The detailed evaluation prompts are provided in Appendix~\ref{appendix:LLM Evaluation Prompts}.

Existing legal AI benchmarks primarily focus on evaluating basic legal knowledge through standardized questions. For instance, LawBench~\cite{fei2023lawbench} assesses models based on legal provision recitation, question answering, and dispute focus identification. However, our experiments reveal that current legal language models, despite their specialized training on such benchmarks, suffer from severe overfitting to static tasks and exhibit significant degradation in dynamic courtroom dialogue. 

To address this limitation, we introduce CourtBench, a courtroom dialogue-focused evaluation dataset comprising 124 multiple-choice questions. CourtBench was constructed by first using GPT-4o-mini to generate questions based on real court case backgrounds, followed by thorough validation and refinement by a team of senior lawyers to ensure quality and practical relevance. Each question presents a comprehensive courtroom dialogue scenario, including case background, prior exchanges between the judge and attorneys, and multiple response options reflecting different legal strategies.

For implementation, we set the temperature parameter to 0.7 for lawyer agents in debates and 0.2 for LLM evaluation to maintain assessment consistency. Each lawyer agent generates approximately 3,361 tokens per round of court debate. All experiments were conducted on a single NVIDIA A100 GPU (80GB), with the simulation of 1,000 cases completing in seven days through the knowledge evolution process.

We compare our GPT-4o-mini-1000 model (which undergoes self-evolution through the simulation of 1,000 cases) against both general-purpose models (GPT-4o-mini as the base architecture and GPT-4o-mini+RAG, which incorporates BGE-M3 embedding over our case database) and specialized legal models, including HanFei-7B~\cite{HanFei} (designed for legal QA and dialogue), LawyerLLaMA-13B~\cite{huang2023lawyer} (fine-tuned on a Chinese legal corpus), and ChatLaw-33B (enhanced with a knowledge graph and mixture-of-experts).

\subsection{Experimental Results}
\begin{figure*}[t]
    \centering
    \begin{subfigure}[b]{0.48\textwidth}
        \centering
        \includegraphics[width=\textwidth]{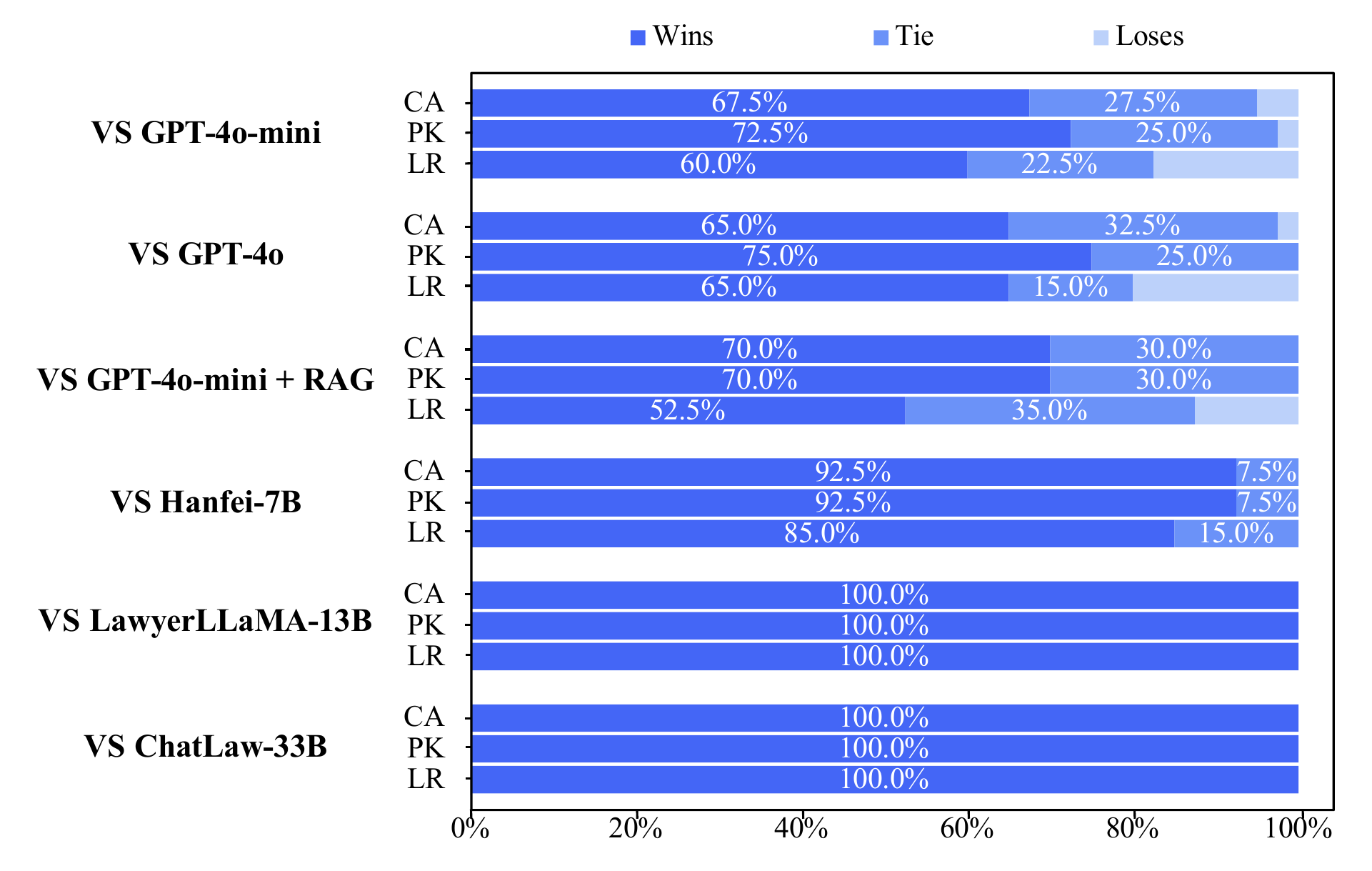}
        \caption{Human evaluation results}
        \label{fig:human_baseline}
    \end{subfigure}
    \hfill
    \begin{subfigure}[b]{0.48\textwidth}
        \centering
        \includegraphics[width=\textwidth]{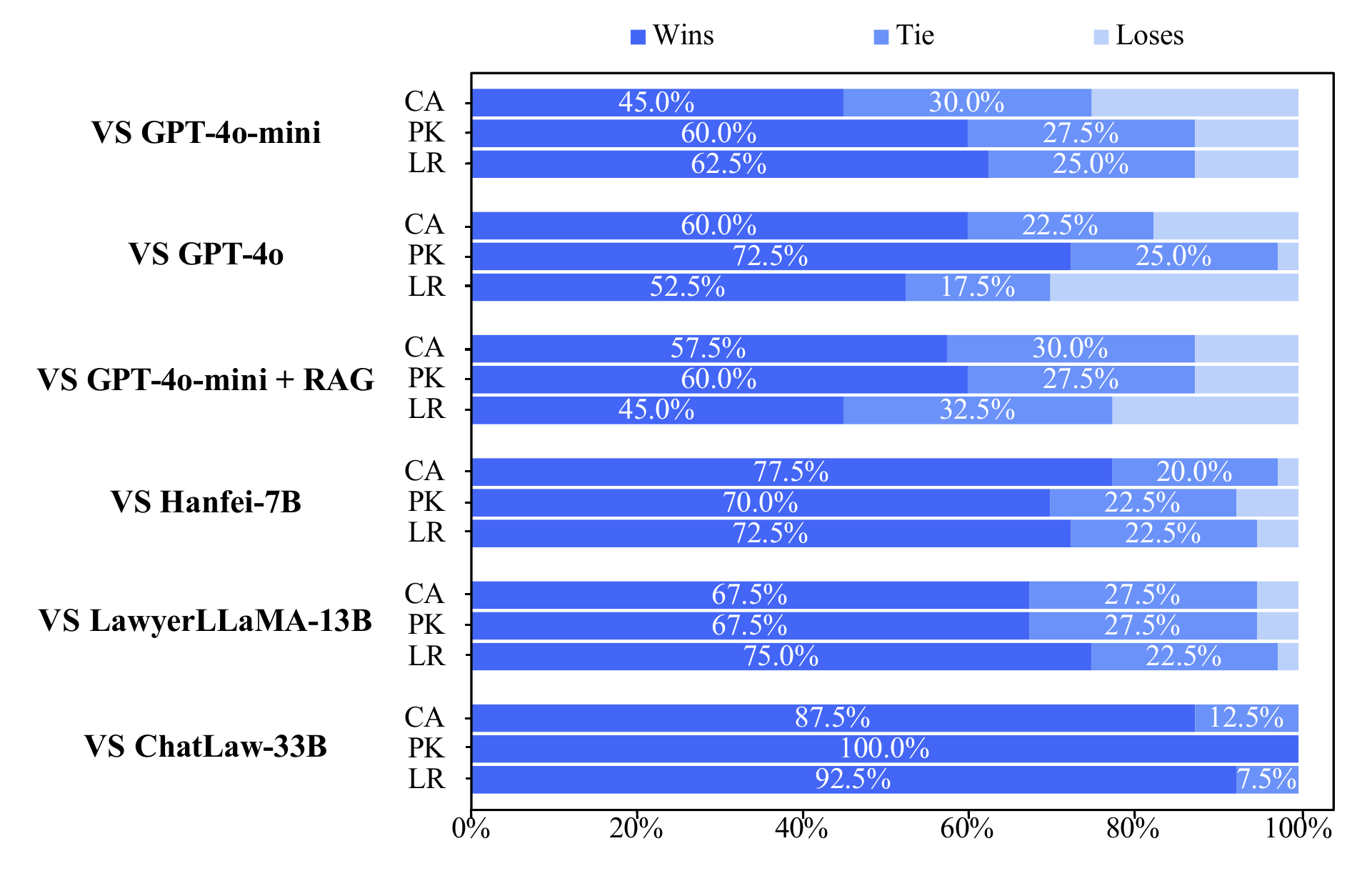}
        \caption{LLM evaluation results}
        \label{fig:llm_baseline}
    \end{subfigure}
\caption{Performance comparison across three dimensions—Cognitive Agility (CA), Professional Knowledge (PK), and Logical Rigor (LR). GPT-4o-mini-1000 consistently outperforms both general-purpose models (GPT-4o-mini, GPT-4o-mini+RAG) and specialized legal models (HanFei-7B, LawyerLLaMA-13B ChatLaw-33B).}
    \label{fig:baseline_comparison}
\end{figure*}

Our comparative analysis highlights the effectiveness of GPT-4o-mini-1000 through systematic evaluation (Figure~\ref{fig:baseline_comparison}). Compared to other GPT-4o-mini variants, our approach demonstrates substantial improvements. Against the retrieval-augmented baseline (GPT-4o-mini+RAG), GPT-4o-mini-1000 consistently achieves higher performance across multiple dimensions, with win rates of 70.0\% in both cognitive agility and professional knowledge in human evaluation. These results suggest that our adversarial evolution approach enables more sophisticated legal reasoning beyond simple retrieval-based enhancements. Furthermore, our model exhibits competitive performance against GPT-4o, achieving a 75.0\% win rate in professional knowledge under human evaluation.

On CourtBench, as shown in Table~\ref{tab:results}, our knowledge-enhanced model achieves performance comparable to GPT-4o (64.52\% vs. 66.13\%), while significantly outperforming the base GPT-4o-mini (52.42\%) and GPT-4o-mini+RAG (58.06\%). This improvement is particularly notable, as it demonstrates that our adversarial evolution approach effectively enhances the base model’s capabilities in dynamic courtroom scenarios without requiring additional model parameters or extensive pre-training.

The experimental results reinforce our observations from Section~\ref{sec:llms_legal} regarding the limitations of specialized legal models in dynamic courtroom interactions. Despite their advanced architectures and domain-specific training, these models exhibit significant performance degradation in interactive settings. This is particularly evident in CourtBench evaluations, where even the largest specialized model, ChatLaw-33B, achieves only 0.81\% accuracy. Similarly, Lawyer-LLaMA-13B and HanFei-7B attain accuracies of 33.87\% and 26.61\%, respectively—substantially lower than their performance on traditional static legal tasks. These findings empirically validate the challenge of adapting models trained primarily on static legal tasks to dynamic courtroom dialogue.

\begin{figure*}[t]
    \centering
    \begin{subfigure}[b]{0.48\textwidth}
        \centering
        \includegraphics[width=\textwidth]{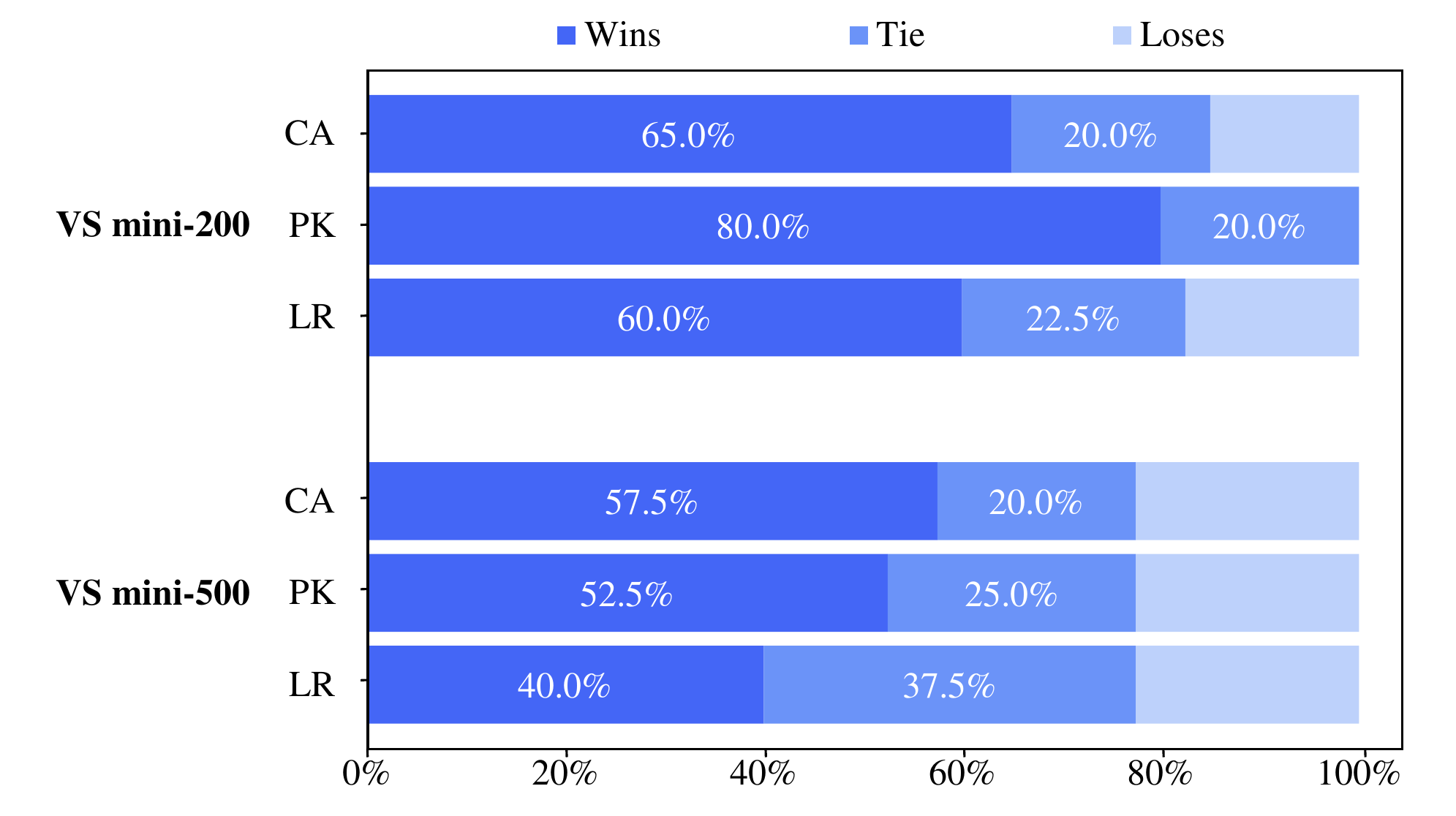}
        \caption{Human evaluation results}
        \label{fig:human_data_size}
    \end{subfigure}
    \hfill
    \begin{subfigure}[b]{0.48\textwidth}
        \centering
        \includegraphics[width=\textwidth]{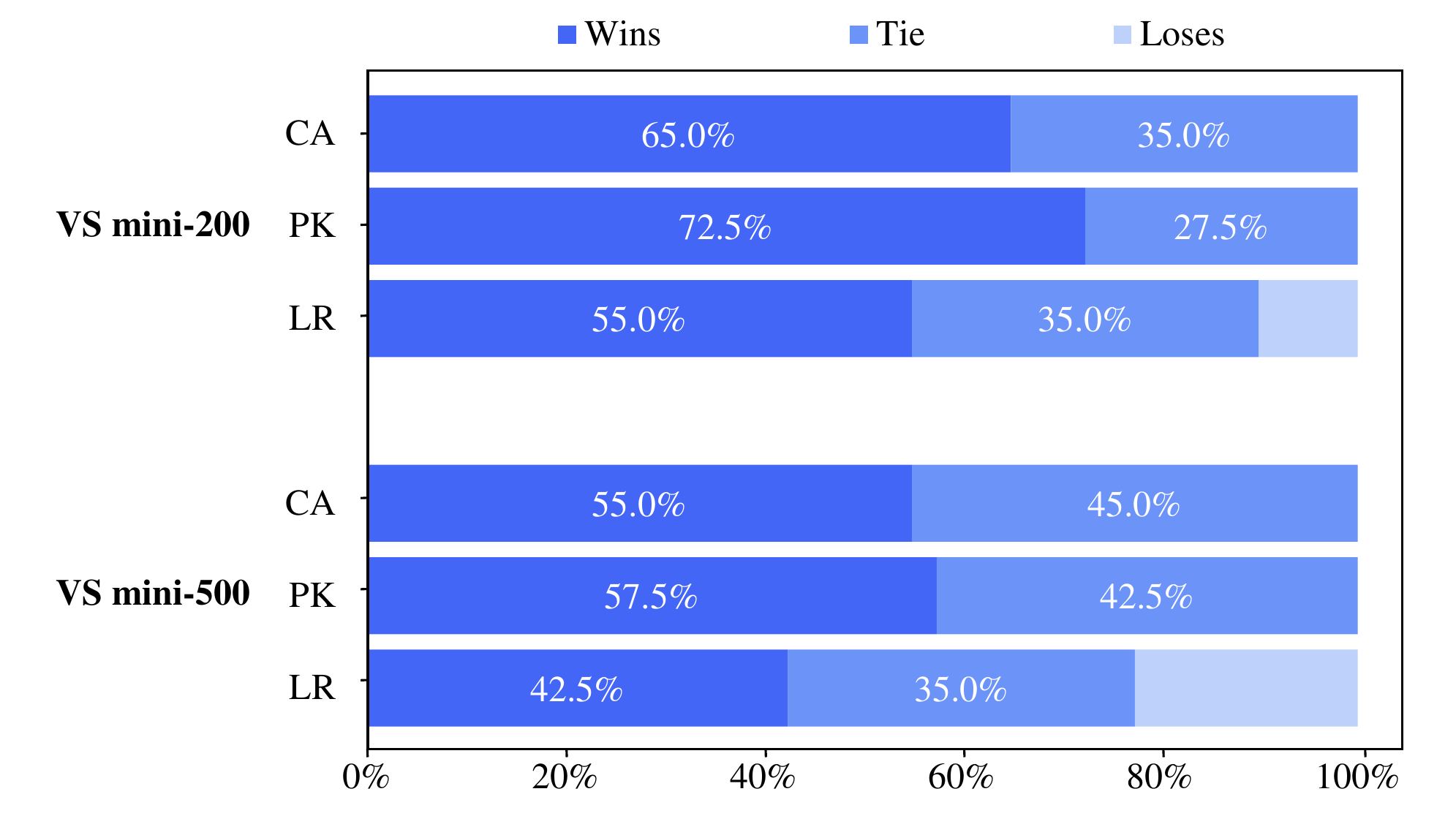}
        \caption{LLM evaluation results}
        \label{fig:llm_data_size}
    \end{subfigure}
    \caption{Impact of training data scale on model performance. Results compare GPT-4o-mini-1000 against models trained on smaller datasets (mini-200, mini-500) across three dimensions—Cognitive Agility (CA), Professional Knowledge (PK), and Logical Rigor (LR)—highlighting the influence of training data size on model capabilities.}
    \label{fig:data_size}
\end{figure*}

Additionally, we investigate the impact of training data scale by comparing GPT-4o-mini-1000 with variants trained on 200 and 500 cases. The comparison reveals distinct scaling patterns across different evaluation dimensions (Figure~\ref{fig:data_size}). In professional knowledge, GPT-4o-mini-1000 demonstrates strong advantages over the 200-case variant (80.0\% and 72.5\% win rates in human and LLM evaluations, respectively). However, for the 500-case variant, the advantage narrows (52.5\% and 57.5\%). This pattern suggests that professional knowledge acquisition initially benefits significantly from increased training data but may reach a plateau.

\begin{table}[t]
\small
\centering
\begin{tabular}{lcc}
\toprule
\textbf{Model} & \textbf{Acc.(\%)} & \textbf{Correct/Total} \\
\midrule
GPT-4o & 66.13 & 82/124 \\
GPT-4o-mini-1000 (Ours) & 64.52 & 80/124 \\
GPT-4o-mini+RAG & 58.06 & 72/124 \\
GPT-4o-mini & 52.42 & 65/124 \\
Lawyer-LLaMA-13B & 33.87 & 42/124 \\
HanFei-7B & 26.61 & 33/124 \\
ChatLaw-33B & 0.81 & 1/124 \\
\bottomrule
\end{tabular}
\caption{Model performance on CourtBench. Our knowledge-enhanced model achieves competitive results with GPT-4o while significantly outperforming specialized legal models.}
\label{tab:results}
\end{table}

\subsection{Ablation Studies}
\begin{figure*}[t]
    \centering
    \begin{subfigure}[b]{0.48\textwidth}
        \centering
        \includegraphics[width=\textwidth]{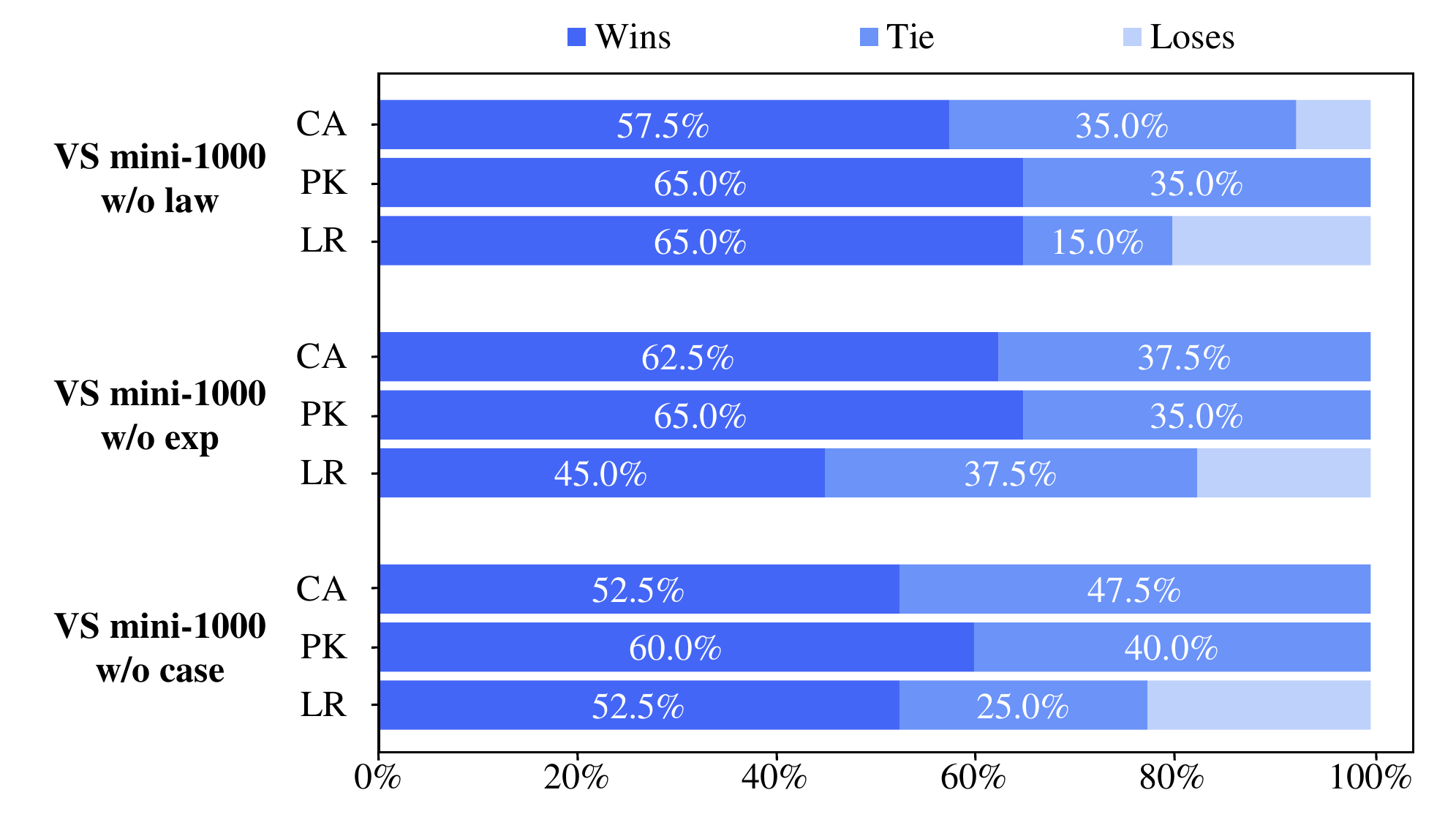}
        \caption{Human evaluation results}
        \label{fig:human_ablation}
    \end{subfigure}
    \hfill
    \begin{subfigure}[b]{0.48\textwidth}
        \centering
        \includegraphics[width=\textwidth]{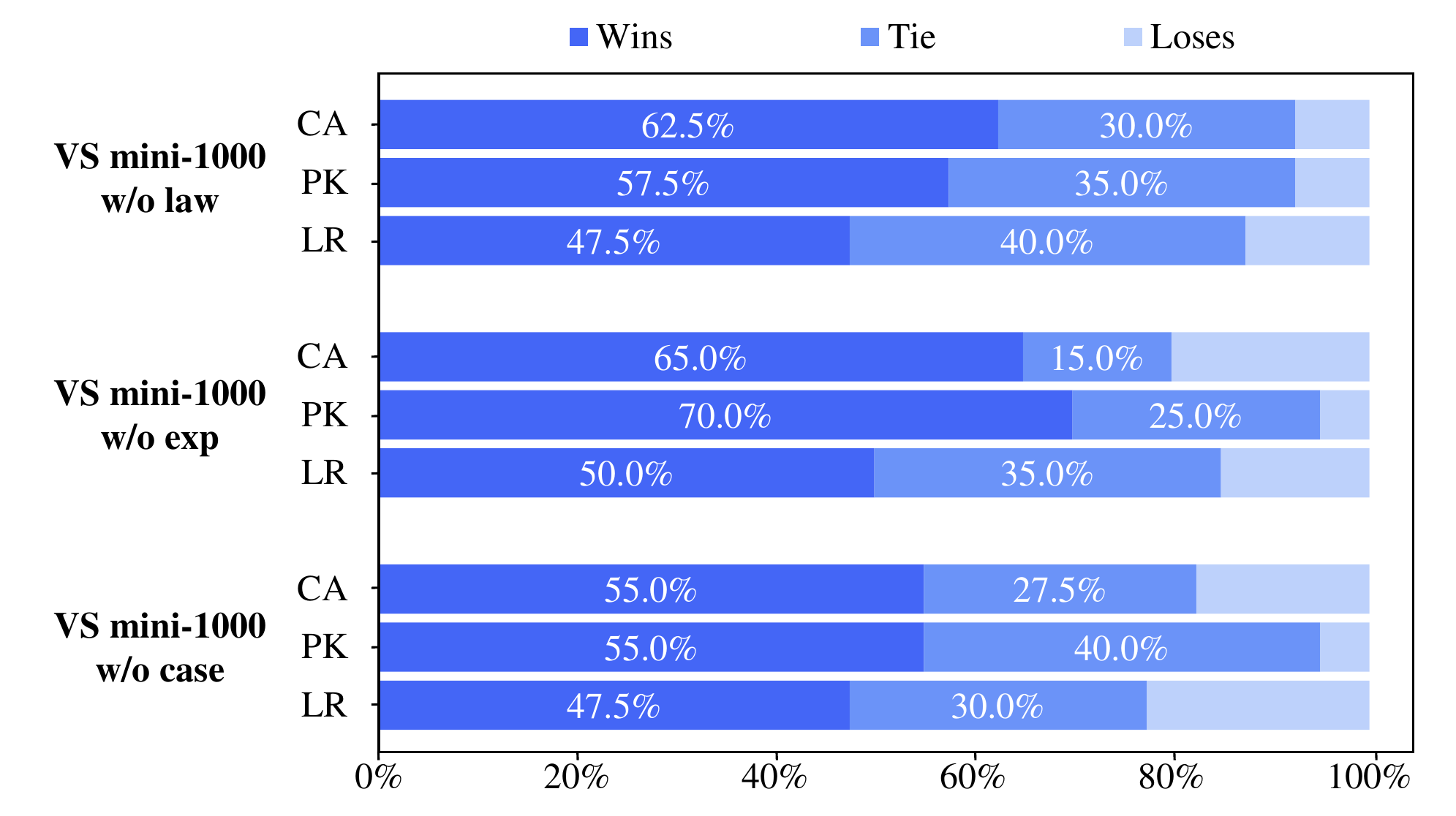}
        \caption{LLM evaluation results}
        \label{fig:llm_ablation}
    \end{subfigure}
    \caption{Ablation study results illustrating the impact of removing different knowledge bases from GPTM-1000. Performance degradation is evaluated by excluding the legal provisions database (w/o law), experience database (w/o exp), and case database (w/o case) across three dimensions: Cognitive Agility (CA), Professional Knowledge (PK), and Logical Rigor (LR).}
        \label{fig:ablation_study}
\end{figure*}

Our ablation experiments reveal distinct contribution patterns of the different knowledge bases (Figure~\ref{fig:ablation_study}). In human evaluation, removing the legal provisions database results in decreased performance across all dimensions, with the model achieving win rates of 57.5\%, 65.0\%, and 65.0\% in cognitive agility, professional knowledge, and logical rigor, respectively. The experience base plays a particularly crucial role in cognitive agility, as its removal reduces the model’s win rate to 62.5\% in human evaluation. The exclusion of case knowledge has a broader impact across all dimensions, with win rates dropping to 52.5\%, 60.0\%, and 52.5\%.

LLM evaluation results exhibit similar patterns, reinforcing the significance of each knowledge base component. Notably, the legal provisions database is essential for maintaining logical rigor, while the experience base has a substantial influence on cognitive agility. These findings empirically validate our integrated knowledge evolution design, demonstrating that each component plays a distinct and essential role in the model’s overall performance.

\subsection{Case Analysis}
To assess our approach’s effectiveness in real-world legal scenarios, we analyze a contract dispute case where GPT-4o-mini-1000 assumes both plaintiff and defendant roles against GPT-4o-mini. The complete case study and additional examples are provided in Appendix~\ref{appendix:Case Study Analysis} (see Figure~\ref{fig:case_study} for a detailed comparison).

Our model effectively integrates the three knowledge bases. In legal knowledge (\textbf{yellow boxes}), it accurately cites relevant provisions, such as Maritime Law Article 57 and Civil Procedure Law Article 57, forming well-structured legal reasoning chains. The experience base enables a systematic five-element argumentation structure and a sophisticated evidence combination strategy, incorporating contracts, bills of lading, and payment records. The case knowledge base enhances professional analysis in shipping agency disputes, particularly regarding unauthorized cargo release.

In contrast, the baseline model (\textbf{blue boxes}) exhibits limited legal reasoning, relying on superficial citations, unsystematic argumentation focused on peripheral issues, and insufficient industry-specific expertise. The judgment results (\textbf{red boxes}) further validate our approach’s effectiveness: as the plaintiff, securing USD 27,509.40 in compensation and litigation costs; as the defendant, successfully defending against all claims. This bidirectional success underscores our model’s balanced capabilities in legal argumentation (detailed in Appendix~\ref{appendix:Case Study Analysis}).

\section{Conclusion}
In this paper, we introduced \textbf{AgentCourt}, a simulation framework for courtroom scenarios that leveraged Large Language Models (LLMs) and agent-based adversarial evolution. Our approach enabled legal agents to dynamically acquire knowledge and refine argumentation strategies, addressing the limitations of traditional static legal AI systems. Through adversarial evolution, agents improved their legal reasoning capabilities by continuously adapting to new case contexts. The integration of a structured three-tier knowledge base, comprising legal provisions, case precedents, and strategic experience, allowed for more comprehensive legal understanding and reasoning. Empirical results demonstrated that GPT-4o-mini-1000 consistently outperformed both general-purpose models and specialized legal models, achieving state-of-the-art performance in dynamic courtroom dialogues. Additionally, we introduced \textbf{CourtBench}, a benchmark designed to evaluate legal AI models based on real courtroom interactions rather than static legal knowledge retrieval, further validating the effectiveness of our approach. While our work provided significant advancements in legal AI, future research could focus on handling more complex legal scenarios, improving role adaptability, and extending the framework to diverse legal systems. By open-sourcing our privacy-anonymized datasets and implementation, we aimed to facilitate further research in this domain.

\section*{Limitations}
Our implementation utilized API calls for agent simulation, significantly reducing memory requirements and computational overhead. However, token generation speed (approximately 3,361 tokens per round) and API response latency remained potential challenges for large-scale applications, particularly in real-time legal advisory systems. While this approach proved more resource-efficient than running full models locally, its reliance on external API calls could introduce variability in response times and potential constraints on long-term knowledge evolution. Additionally, our framework primarily focused on adversarial learning within a structured legal setting, which may require further adaptations to generalize effectively across diverse legal systems and jurisdictions. Future work could explore optimization strategies to enhance inference speed, as well as techniques to improve model adaptability in cross-jurisdictional legal applications.

\section*{Ethics Statement}
All civil court cases used in our study were obtained from publicly accessible sources, with sensitive information properly anonymized to protect privacy. AgentCourt is designed as a training and research tool to enhance legal professionals' capabilities and advance our understanding of legal AI systems. We acknowledge several important ethical considerations. First, while AgentCourt demonstrates promising results in simulated court proceedings, it is not intended to replace human legal professionals or make actual legal decisions. The system should be used as a supplementary tool for legal training and research purposes only. Second, the legal knowledge and strategies learned by our system should not be misused for generating deceptive legal arguments or manipulating court proceedings. Furthermore, we emphasize that the outputs generated by AgentCourt require careful review and validation by qualified legal professionals before any practical application. The system's responses should not be considered as formal legal advice or used directly in real court proceedings without proper human oversight.

\section*{Acknowledgments}
This work was partially supported by National Key Research and Development Program of China (2024YFF0908200), National Natural Science Foundation of China (62376262), the Natural Science Foundation of Guangdong Province of China (2025B1515020032, 2024A1515030166), GuangDong Basic and Applied Basic Research Foundation (2023A1515110718 and 2024A1515012003), China Postdoctoral Science Foundation (2024M753398), Postdoctoral Fellowship Program of CPSF (GZC20232873).
\bibliography{custom}

\appendix

\section{System Overview}
\label{appendix:System Overview}
\begin{figure*}[h]
  \centering
  \includegraphics[width=1.0\linewidth]{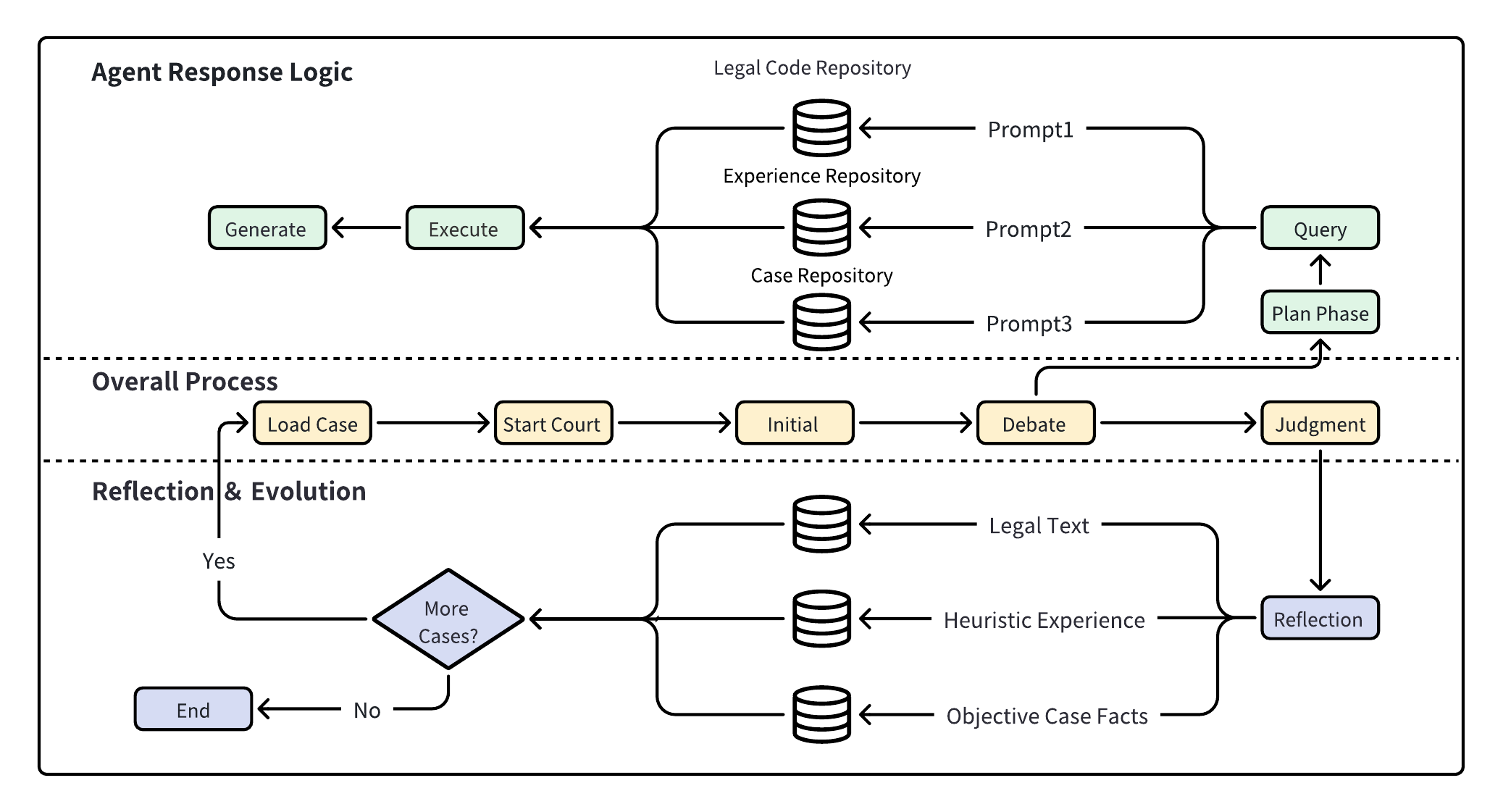}
  \caption{Simulation of the court process. This figure illustrates the complete workflow of the simulated court: (1) The middle row outlines the overall court framework; (2) During the free debate phase, each agent retrieves relevant knowledge from the three databases as needed to enhance their responses; (3) Upon completing a case simulation, the agent reflects and evolves, continuously expanding its knowledge bases.}
  \label{fig:overall process}
\end{figure*}

Figure~\ref{fig:overall process} provides a comprehensive overview of the \textbf{AgentCourt} system framework and workflow.

\section{Agents and Prompts}
\label{appendix:Agents and Prompts}
\subsection{Agent Visualization}
\label{appendix:Agent Visualization}
\begin{figure}[h]
  \centering
  \includegraphics[width=1\linewidth]{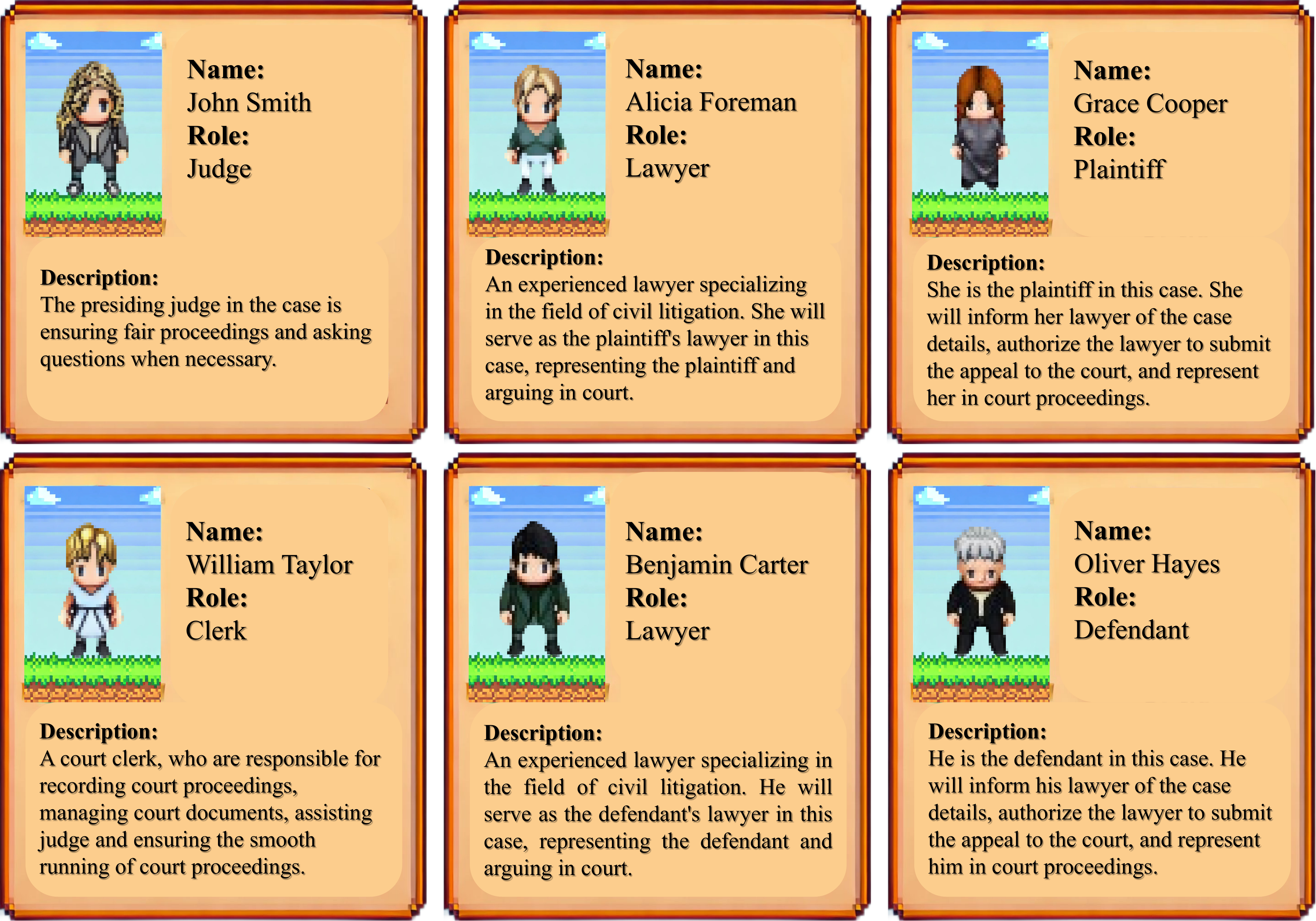}
  \caption{Example agents in \textbf{AgentCourt}.}
  \label{fig:agent}
\end{figure}

As illustrated in Figure~\ref{fig:agent}, each agent is assigned specific roles and responsibilities within the court simulation process.

\subsection{Lawyer Agent Prompts}
\label{appendix:Lawyer Agent Prompts}
This section presents the detailed prompt templates used for lawyer agents in our system. As shown in Table~\ref{tab:lawyer_prompts}, the lawyer agent first determines the required information sources using a structured prompt template. Tables~\ref{tab:legal_reflection}, \ref{tab:experience_reflection}, and \ref{tab:case_reflection} illustrate the reflection prompts designed for different knowledge bases: legal provisions, experience, and case library, respectively. These structured prompts ensure the lawyer agents exhibit consistent and professional behavior throughout the court proceedings.

\begin{table*}[t]
\centering
\begin{tabular}{|p{0.12\textwidth}|p{0.8\textwidth}|}
\hline
\textbf{Phase} & \textbf{Prompt Template} \\
\hline
Information Planning & \texttt{You are a [ROLE]. [DESCRIPTION] As an experienced legal professional, analyze the court history and determine required information sources.}\newline
Return format: \{\texttt{'experience': bool, 'case': bool, 'legal': bool}\} \\
\hline
Experience Query & \texttt{Based on the court history, analyze required experience information. Identify key points and formulate a query to retrieve relevant experiences for improving logic.}\newline
Return format: \{\texttt{'query': 'specific query string'}\} \\
\hline
Case Query & \texttt{Based on the court history, analyze required case precedents. Identify key points and formulate a query to retrieve relevant cases for improving agility.}\newline
Return format: \{\texttt{'query': 'specific query string'}\} \\
\hline
Legal Query & \texttt{Based on the court history, analyze required legal information. Identify relevant laws/regulations and formulate a query to retrieve legal references.}\newline
Return format: \{\texttt{'query': 'specific query string'}\} \\
\hline
Response Generation & \texttt{Guidelines:}\newline
1. Avoid repeating previous arguments\newline
2. Build upon previous points with new perspectives\newline
3. Respond directly to opponent's latest arguments\newline
4. Introduce new supporting evidence when possible\newline
5. Vary expression and argumentation approach \\
\hline
\end{tabular}
\caption{Prompt templates for lawyer agents in the planning and execution phase. Each prompt is preceded by the basic instruction ``You are a [ROLE]. [DESCRIPTION]''.}
\label{tab:lawyer_prompts}
\end{table*}

\begin{table*}[t]
\centering
\begin{tabular}{|p{0.15\textwidth}|p{0.8\textwidth}|}
\hline
\multicolumn{2}{|c|}{\textbf{Legal Provision Database Reflection}} \\
\hline
Law Usage Analysis & \texttt{Extract all law citations from the response, including:}\newline
1. Explicit legal provisions (e.g., Article X)\newline
2. Implicit legal references\newline
3. Specific legal clauses mentioned\newline
Return format: \{\texttt{'laws': [\{'content', 'purpose', 'issue', 'source'\}]}\} \\
\hline
Effectiveness Evaluation & \texttt{Evaluate the effectiveness of legal provision usage:}\newline
1. Relevance to the issue\newline
2. Persuasiveness of argumentation\newline
3. Effectiveness of opponent's response\newline
4. Overall impact\newline
Return format: \{\texttt{'relevance\_score', 'persuasiveness\_score', 'response\_effectiveness', 'overall\_effectiveness', 'analysis', 'improvement\_suggestions'}\} \\
\hline
Opponent Analysis & \texttt{Analyze opponent's legal provision usage and extract:}\newline
- Law content\newline
- Usage method\newline
- Application effectiveness\newline
Return format: \{\texttt{'laws': [\{'content', 'usage\_method', 'effectiveness'\}]}\} \\
\hline
Opponent Evaluation & \texttt{Evaluate opponent's excellence in:}\newline
1. Professional law selection\newline
2. Logical argumentation\newline
3. Expression techniques\newline
Return format: \{\texttt{'professionalism\_score', 'logic\_score', 'expression\_score', 'overall\_score', 'learning\_points', 'applicable\_scenarios'}\} \\
\hline
\end{tabular}
\caption{Prompt templates for legal provision database reflection phase.}
\label{tab:legal_reflection}
\end{table*}

\begin{table*}[t]
\centering
\begin{tabular}{|p{0.15\textwidth}|p{0.8\textwidth}|}
\hline
\multicolumn{2}{|c|}{\textbf{Experience Database Reflection}} \\
\hline
Experience Summary & \texttt{Generate a logically coherent experience summary including:}\newline
1. Case background description\newline
2. Logic-focused experience description\newline
3. 3-5 key points for practical application\newline
4. 3-5 guidelines for maintaining logical coherence\newline
Return format: \{\texttt{'context', 'content', 'focus\_points', 'guidelines'}\} \\
\hline
Learning Experience & \texttt{Summarize learnings from opponent's performance:}\newline
1. Legal provision application techniques\newline
2. Argumentation construction methods\newline
3. Persuasive expression points \\
\hline
\end{tabular}
\caption{Prompt templates for experience database reflection phase.}
\label{tab:experience_reflection}
\end{table*}

\begin{table*}[t]
\centering
\begin{tabular}{|p{0.15\textwidth}|p{0.8\textwidth}|}
\hline
\multicolumn{2}{|c|}{\textbf{Case Library Reflection}} \\
\hline
Case Summary & \texttt{Generate a concise case summary for agile response:}\newline
1. Case name and background\newline
2. Case type (e.g., labor dispute, contract dispute)\newline
3. 3-5 essential keywords\newline
4. 3-5 quick reaction points\newline
5. 3-5 response directions\newline
Return format: \{\texttt{'content', 'case\_type', 'keywords', 'quick\_reaction\_points', 'response\_directions'}\} \\
\hline
\end{tabular}
\caption{Prompt templates for case library reflection phase.}
\label{tab:case_reflection}
\end{table*}

\subsection{Judge Agent Prompts}
\label{appendix:Judge Agent Prompts}
The judge agent plays a crucial role in our court simulation system. Table \ref{tab:judge_prompts} presents the prompt templates used for the judge agent.
\begin{table*}[t]
\centering
\begin{tabular}{|p{0.15\textwidth}|p{0.8\textwidth}|}
\hline
\multicolumn{2}{|c|}{\textbf{Judge Agent Base Configuration}} \\
\hline
Role Description & \texttt{You are the presiding judge in this case, responsible for conducting the trial, ensuring procedural fairness, and raising questions when necessary.} \\
\hline
\hline
\multicolumn{2}{|c|}{\textbf{Judge Agent Core Functions}} \\
\hline
Initial Question & \texttt{Based on the statements from both plaintiff's and defendant's attorneys, summarize the key points for debate. Your summary should be concise and practical while adhering to reality.} \\
\hline
Final Judgment & \texttt{Please make your judgment: (Your decision should align with realistic circumstances.)} \\
\hline
\end{tabular}
\caption{Prompt templates for the judge agent.}
\label{tab:judge_prompts}
\end{table*}

\section{Simulation Workflow Details}
\label{appendix:Simulation Workflow Details}
The system operation consisted of three phases to facilitate comprehensive legal knowledge acquisition: pre-trial preparation, court proceedings, and knowledge construction.

In the pre-trial phase, we constructed an experimental dataset using 1,000 real civil cases from courts in Beijing, Shanghai, and Shenzhen between 2018 and 2020. Complaints and defense statements were extracted as simulation backgrounds, with detailed data processing outlined in Appendix~\ref{appendix:Data Settings and Processing}. To ensure diverse learning experiences, we implemented a random role assignment mechanism, allowing lawyer agents to gain multi-dimensional expertise from different litigation perspectives.

The court proceedings phase followed standard judicial procedures. The simulation began with the clerk announcing court discipline to establish trial order, followed by the judge formally opening the session and verifying the identities and qualifications of the litigation participants. During the case presentation phase, lawyers for both the plaintiff and defendant presented their respective claims and defenses. Based on these statements, the judge identified and summarized key disputed points to structure the subsequent debates. The debate phase involved multiple rounds of argumentation, where lawyer agents leveraged their knowledge bases to strengthen their reasoning and counterarguments. Finally, the judge rendered a ruling based on a comprehensive assessment of the case, and the clerk completed the trial records for archiving.

In the knowledge construction phase, lawyers from both sides reflected on the court session, focusing on enhancing their legal provision capabilities, conducting self-reflection, learning from opponents, and refining case analysis. This iterative learning process enabled the continuous evolution of their debate strategies, leading to improved performance in subsequent cases.

\section{Data Settings and Processing}
\label{appendix:Data Settings and Processing}
\begin{figure*}[h]
  \centering
  \includegraphics[width=1.0\linewidth]{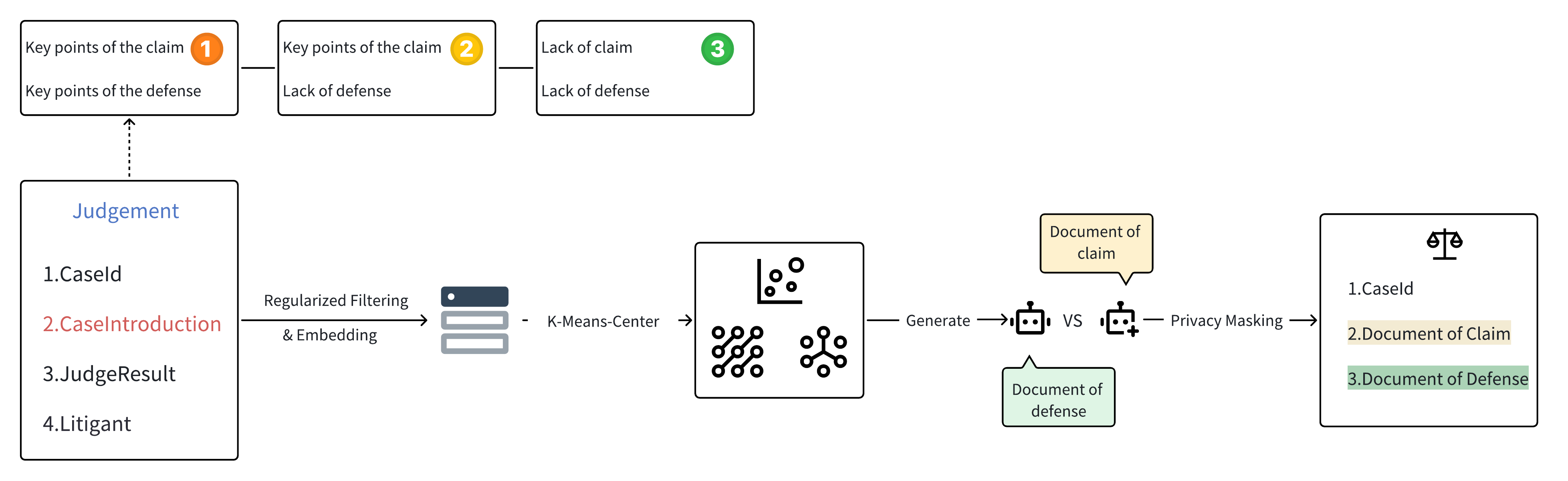}
  \caption{Data processing workflow. The initial collection of legal documents can be categorized into three types as labeled by numbers 1-3 in the figure: (1) documents containing both indictment and defense information, (2) documents with only indictment information (lacking defense content), and (3) documents lacking both indictment and defense information. We ultimately adopted the first category to extract structured indictment petitions and defense statements.}
  \label{fig:data_setting_flows}
\end{figure*}

Our data processing pipeline, as illustrated in Figure~\ref{fig:data_setting_flows}, encompasses regularized filtering, BERT-based embedding, and privacy masking.

\subsection{Accessing Confidential Pleadings}
Pleadings play a fundamental role in legal proceedings but are often confidential and proprietary, limiting access to these critical documents. Traditional open-source datasets are insufficient, as primary case files are typically restricted within court filing systems and private legal records.

\subsection{Dataset Construction and Preprocessing}
Using the China Judgement Website\footnote{https://wenshu.court.gov.cn/}, we compiled a dataset of 10,000 civil judgments. To enhance dataset quality, we applied meticulous preprocessing steps, selecting 1,389 high-value cases that included both plaintiff claims and defendant defenses. To mitigate redundancy, we employed BERT-based semantic vectorization \cite{cui2021pre} on the 'Case Introduction' sections and applied K-Means clustering \cite{kodinariya2013review} to group similar documents. This process resulted in a refined dataset of 1,000 representative cases used for our moot court training and evaluation.

\subsection{Data Generation and Anonymization}
We leveraged the ERNIE-Speed-128K API \cite{ERNIE-Speed-128K} to generate and anonymize high-fidelity simulated legal texts tailored to civil judgments. This yielded a curated dataset comprising 1,000 training samples and 50 test samples, facilitating robust legal argumentation and judgment prediction within our simulated moot court environment. This structured approach enhances the reliability of legal AI models while ensuring data privacy compliance.

\section{Case Study Analysis}
\label{appendix:Case Study Analysis}
\begin{figure*}[t]
\centering
\begin{subfigure}{\textwidth}
\includegraphics[width=\textwidth]{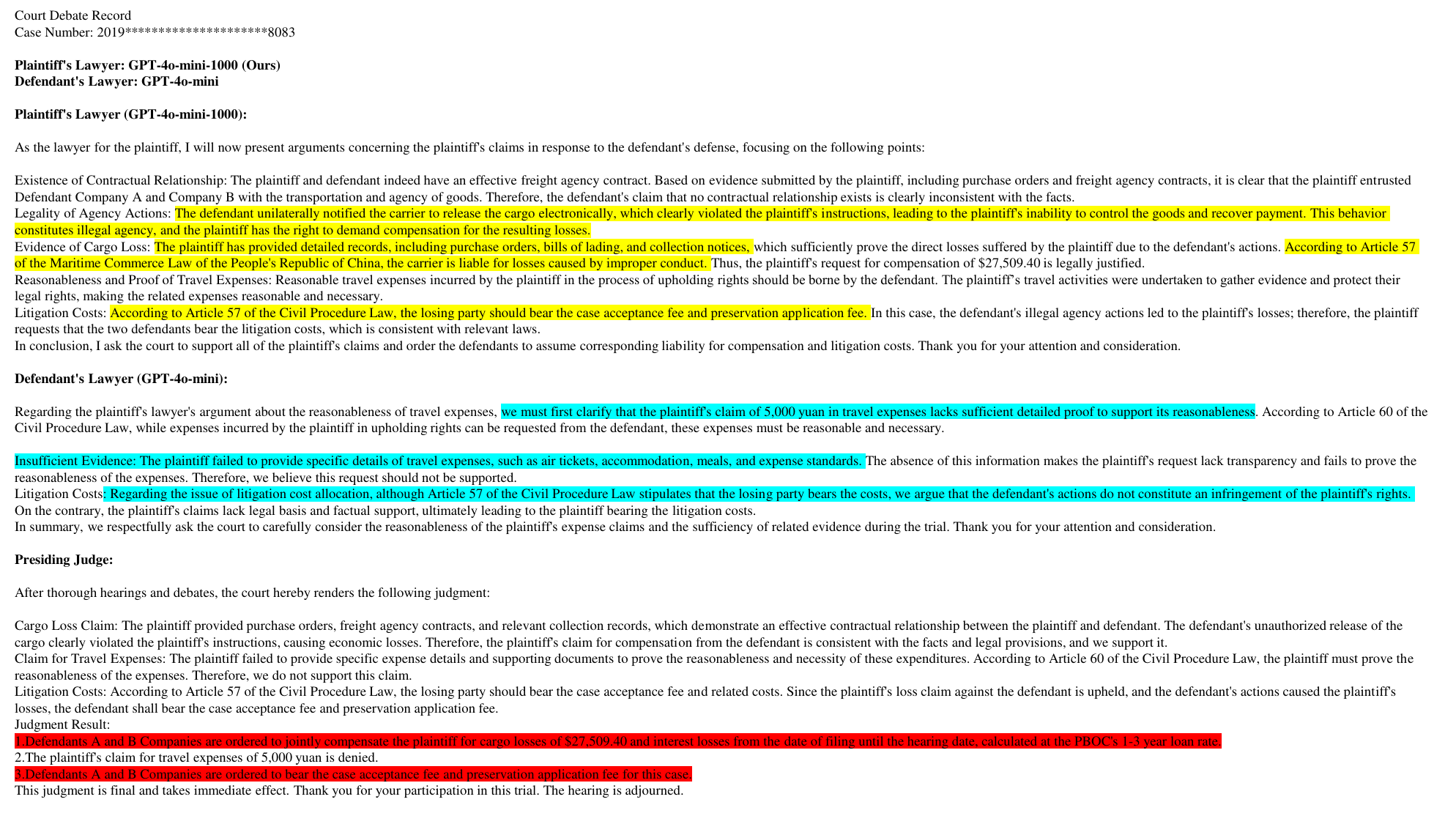}
\caption{GPT-4o-mini-1000 as plaintiff vs GPT-4o-mini as defendant}
\label{fig:case_study_plaintiff}
\end{subfigure}

\begin{subfigure}{\textwidth}
\includegraphics[width=\textwidth]{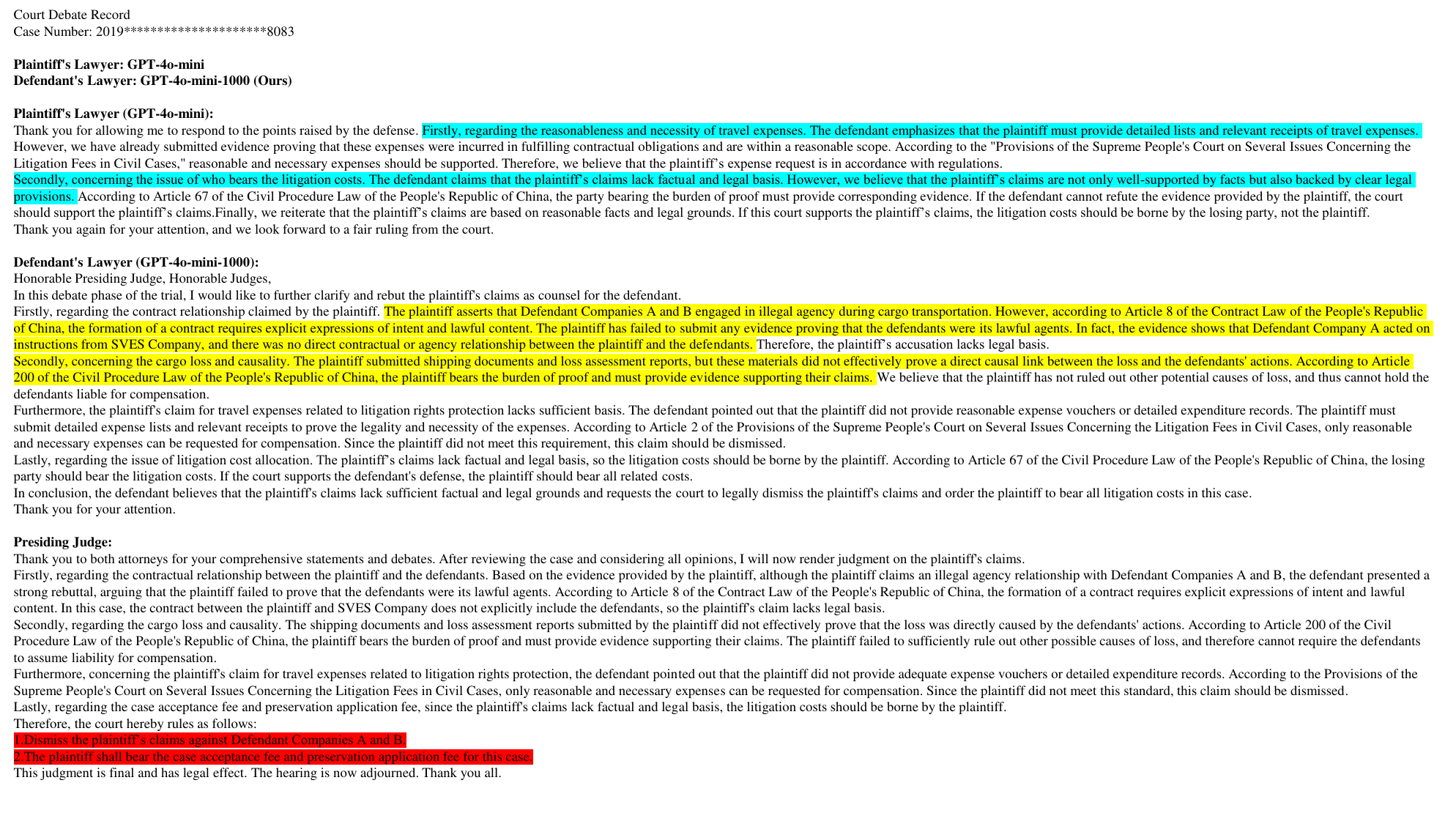}
\caption{GPT-4o-mini-1000 as defendant vs GPT-4o-mini as plaintiff}
\label{fig:case_study_defendant}
\end{subfigure}

\caption{Comparison of court debate performance in a contract dispute case. \textbf{Yellow boxes} highlight GPT-4o-mini-1000's effective use of knowledge bases, including precise legal citations, systematic argument structures, and professional domain expertise. \textbf{Blue boxes} indicate GPT-4o-mini's deficiencies, such as superficial arguments and a lack of professional depth. \textbf{Red boxes} emphasize the contrasting judgment outcomes, where GPT-4o-mini-1000 secures compensation as the plaintiff and successfully defends claims as the defendant. Due to the length of the full court records, both figures present excerpts from the most representative round of debates.}

\label{fig:case_study}
\end{figure*}

To evaluate the effectiveness of our approach, we present a detailed case study analysis in Figure~\ref{fig:case_study}. We compare the performance of GPT-4o-mini-1000 against the baseline model in both plaintiff and defendant roles, as illustrated in Figure~\ref{fig:case_study_plaintiff} and Figure~\ref{fig:case_study_defendant}.

\section{Evaluation Prompts}
\label{appendix:Evaluation Prompts}

\subsection{LLM Evaluation Prompts}
\label{appendix:LLM Evaluation Prompts}
For automatic evaluation of court debates, we designed structured prompts to guide the LLM evaluator. As shown in Table~\ref{tab:evaluation_prompts}, these prompts assess performance across three key dimensions: cognitive agility, professional knowledge, and logical rigor. Each dimension includes specific criteria and scoring guidelines to ensure consistent and objective evaluation.

\subsection{CourtBench Assessment Prompts}
\label{appendix:CourtBench Assessment Prompts}
The \textbf{CourtBench} dataset evaluation employs two types of prompts, as outlined in Table~\ref{tab:evaluation_prompts}. The base prompt provides standard case evaluation instructions, while the enhanced prompt incorporates additional professional knowledge references for models equipped with external knowledge bases.

\begin{table*}[t]
\centering
\begin{tabular}{|p{0.15\textwidth}|p{0.8\textwidth}|}
\hline
\multicolumn{2}{|c|}{\textbf{Automatic Evaluation Prompts}} \\
\hline
\textbf{Debate Evaluation} & \texttt{As a senior legal expert, please evaluate the following court debate across three dimensions:}\newline
1. Cognitive Agility: depth of understanding, response speed, and accuracy\newline
2. Professional Knowledge: expertise in legal knowledge and application\newline
3. Logical Rigor: completeness of argumentation and reasoning structure\newline
\textbf{Return format:} \{\texttt{'cognitive\_agility': 1/0/-1, 'professional\_knowledge': 1/0/-1, 'logical\_rigor': 1/0/-1, 'overall': 1/0/-1}\}\newline
where 1 = plaintiff wins, 0 = tie, -1 = defendant wins. \\
\hline
\multicolumn{2}{|c|}{\textbf{CourtBench Dataset Evaluation}} \\
\hline
\textbf{Base Prompt} & \texttt{As a senior legal expert, please select the most appropriate answer based on the following case information:}\newline
- Case Background: [background]\newline
- Court Process: [court\_process]\newline
- Current Focus: [focus]\newline
- Question: [question]\newline
- Options: A/B/C/D\newline
\textbf{Return format:} Single letter (A/B/C/D). \\
\hline
\textbf{Enhanced Prompt} & Base prompt + \texttt{Professional Knowledge Reference: [reference]}\newline
(Used for models with external knowledge bases, e.g., GPT-4o-mini-1000 and GPT-4o-mini-RAG). \\
\hline
\end{tabular}
\caption{Evaluation prompts for automated assessment and CourtBench dataset evaluation.}
\label{tab:evaluation_prompts}
\end{table*}

\section{LawBench Result}
\label{appendix:LawBench Result}
As shown in Table~\ref{tab:legal-tasks}, legal-specific models demonstrate decent performance on the LawBench dataset. However, the tasks in these tests are mostly specific legal knowledge questions, such as reciting specific laws or identifying legal dispute focuses. In the legal field, courtroom debate ability is a crucial component, which requires not only the ability to recite laws but also the ability to apply them reasonably to assist in debates. It can be observed that these models show a decline in performance on our CourtBench dataset.
\begin{table*}[t]
\small
\centering
\begin{tabular}{lcccc}
\toprule
\textbf{Model} & \textbf{MEM (\%)} & \textbf{UND1 (\%)} & \textbf{UND2 (\%)} & \textbf{APP (\%)} \\
\midrule
GPT-4o-mini & 16.84 & 18.4 & 31.2 & 16.74 \\
GPT-4o-mini-1000 (Ours) & \textbf{20.2} & \textbf{21.6} & \textbf{33.8} & \textbf{18.34} \\
HanFei-7B & 17.03 & 6.39 & 30.20 & 16.06 \\
Lawyer-LLaMA-13B & 12.33 & 8.25 & 4.40 & 16.94 \\
ChatLaw-33B & 11.74 & 8.04 & 19.80 & 16.55 \\
\bottomrule
\end{tabular}
\caption{Legal Knowledge Tasks Performance. MEM: Memorization Task (Article Recitation, Rouge-L); UND1: Understanding Task1 (Dispute Focus Identification, F1); UND2: Understanding Task2 (Issue Topic Identification, Acc); APP: Applying Task (Consultation, Rouge-L).}
\label{tab:legal-tasks}
\end{table*}
\end{document}